\documentclass[10pt,twocolumn,letterpaper]{article}

\usepackage[pagenumbers]{cvpr} 
\usepackage{stfloats}
\definecolor{cvprblue}{rgb}{0.21,0.49,0.74}
\usepackage[pagebackref,breaklinks,colorlinks,allcolors=cvprblue]{hyperref}
\usepackage{xcolor}
\usepackage{amsthm}   
\newtheorem{hypothesis}{Hypothesis}

\def\confName{CVPR}
\def\confYear{2026}

\title{Robot Generalization Beyond Volume: The Diversity Path via Zero-Shot Capabilities}

\author{Rafael Felix\\
runibiLABS\\
{\tt\small rafa.felix@runibi.ai}
\and
Flavia Yanase\\
runibiLABS\\
{\tt\small flavia.yanase@runibi.ai}
}

\begin{document}
\twocolumn[{%
\renewcommand\twocolumn[1][]{#1}%
\maketitle
    \includegraphics[width=.98\linewidth]{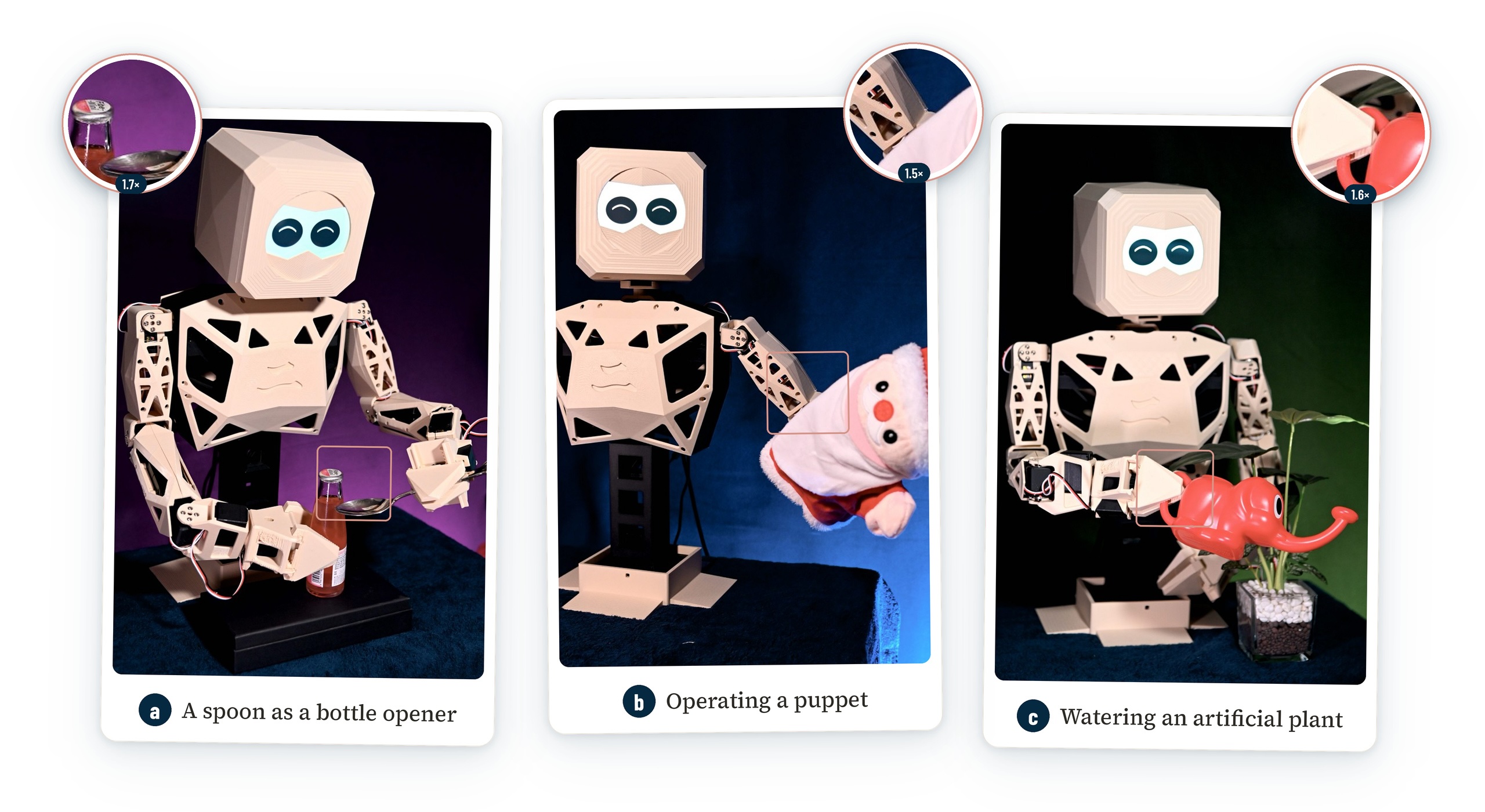}
    \captionof{figure}{Examples of zero-shot generalization, shown on our dual-arm prototype. (a) Opening a bottle with a spoon: everyday human know-how that is rarely captured in images and absent from robot data. (b) Operating a puppet: expert skill with no robot data. (c) An elephant-shaped watering can and an artificial plant: a long-tail object combined with a semantic judgment, since the right outcome is to recognize that the plant needs no water. No single team can anticipate and collect such cases; a diverse community of developers brings them within reach.}

\label{fig:hero}
}]


\begin{abstract}
General-purpose robots must handle tasks, objects, and environments they were never trained on, yet zero-shot generalization remains one of the least solved problems in physical AI. In digital AI, zero-shot capability grew with the volume and, above all, the diversity of training data. Robotics instead relies on paid collection by isolated teams, yielding volume without the diversity needed for open-world deployment. We formulate zero-shot generalization in physical AI as coverage of the deployment distribution, proving that repeated sampling from existing collectors shrinks the unseen region only slowly, whereas independent contributors expand its boundary. Building on these foundations, we introduce a shared latent space for robot actions that unites multi-modal alignment, generative augmentation, and noisy-label learning. To feed this latent space at scale, we establish a decentralized data-sourcing model driven by developers deploying real-world applications on a standardized, low-cost hardware platform: Origami. We conclude with a testable scaling hypothesis and an empirical evaluation protocol directly linking real-robot zero-shot success to the diversity of contributed data.
\end{abstract}

\section{Introduction}
\label{sec:intro}

In recent years, several reports have projected strong growth for robotics through 2050. McKinsey estimates a \$370B general-purpose robotics market by 2040 \cite{mckinsey2025embodied}, Japan has set a national target of 10 million AI robots by 2040 \cite{nhk2026robots}, and Morgan Stanley projects a \$5T humanoid market by 2050 \cite{morganstanley2025humanoids}. These projections are driven by falling hardware costs, growing AI capabilities, and adoption that starts with developers and spreads through society. However, to achieve the true return on investment in physical AI, robot generalization is a major hurdle, and existing research has not yet found scalable solutions \cite{zhou2025exploring, batra2025zero}.


The promise of general-purpose robotics is simple: a user hands a home robot a completely unseen object, and the robot instantly knows how to use it, as illustrated in Figure~\ref{fig:hero}. A general autonomous system should be capable of inferring the new object's affordances, selecting a suitable grasp, and completing tasks with it~\cite{ahn2022saycan, kim2024openvla}. Current robotic systems perform poorly at this capability, known as zero-shot generalization \cite{zhou2025exploring,batra2025zero}. Zero-shot generalization has been the target of extensive studies over the past decades in the field of digital AI such as computer vision \cite{lampert2009learning, xian2019zero}, large language models \cite{radford2019language, brown2020language}, audio \cite{radford2023robust} and multi-modal learning \cite{felix2018multi,felix2020augmentation,felix2019ensemble}. This capability determines whether robots remain confined to structured settings or unlock scalable economic value.

In digital AI, zero-shot generalization has progressed as a function of the volume and diversity of data~\cite{fang2022data,felix2018multi}. Figure~\ref{fig:density-diversity} illustrates that volume is the density with which each situation is represented in the latent space, and diversity is the footprint those situations cover. This mirrors a familiar result for Gaussian processes: posterior uncertainty shrinks only near observed inputs, so spreading samples across the input space reduces uncertainty far more than repeating them in one region~\cite{seeger2004gaussian}. Volume helps AI systems recognize patterns accurately, but diversity allows them to interpolate within the latent space and act on concepts they have never seen. Seminal research and industry efforts have demonstrated zero-shot capabilities in large language models (LLMs) \cite{brown2020language, kojima2022large}, computer vision \cite{radford2021learning}, and other fields \cite{radford2023robust, girdhar2023imagebind}. Robotics has yet to show comparable zero-shot capability. Moreover, the problem of diversity has been largely overlooked, as industry and research efforts have prioritized demonstrations of robots performing single tasks~\cite{chi2023diffusion, zhao2023learning}. Industrial and early-stage robots are typically deployed for a single, repetitive task, and each new task requires reprogramming or new training data \cite{black2024pi0, zhou2025exploring}. A robot able to perform an array of tasks within the same facility, adapting to new parts, layouts and procedures, would amortize its cost across many functions and lower the integration cost of each new task. For instance, in a warehouse, a robot trained to pick a given type of product could take on new types of object, or develop new tasks. In elderly care, a robot that delivers meals could assist carers with preparing food, a new task with new handling demands~\cite{srivastava2021behavior}. Further examples are depicted in Figure~\ref{fig:hero},

We introduce a scalable framework that solves physical AI's data bottleneck by combining zero-shot learning with crowd-sourced data collection. Our approach also leverages generative AI, learning from noisy labels and scalable trust and safety \cite{felix2018multi,garg2023instance,felix2021dicta}. Robotic generalization will emerge from structured combination of decentralized, and long-tail data acquisition. This allows the acquisition of large volumes of data safely, while widening the range of scenarios, objects, situations and environments far beyond what any single engineering team could cover. Unlike digital AI, robots must also transfer what they learn in simulation to the physical world, where simulation never fully matches reality, a challenge known as the sim-to-real gap \cite{zhao2020sim, tobin2017domain}. A decentralized community running the same robotic platform in both simulation and reality exposes that gap at scale, and provides the tools to close it.


The remainder of this paper is organized as follows: Section~\ref{sec:related_work} synthesizes the foundations of zero-shot transfer, latent spaces, and imperfect data sourcing. Section~\ref{sec:method} details our core approach—formalizing distribution coverage, outlining our multi-modal latent space, and introducing the \textit{Real2Sim2Real} platform alongside our distributed data model. Section~\ref{sec:hypothesis} defines our empirical evaluation protocol and scaling hypothesis, followed by limitations (Section~\ref{sec:limitations}) and concluding implications for physical AI (Section~\ref{sec:conclusions}).   


\section{Related Work}
\label{sec:related_work}

This section reviews the topics that form the foundations of our approach: zero-shot capabilities, latent spaces, scaling laws, trust and safety and imperfect crowd-sourced data.


\subsection{Zero-shot capabilities}
\label{sec:rw/zero-shot-capabilities}

Zero-shot generalization is the defining bottleneck preventing autonomous robots from operating beyond structured factory floors. Large-scale web platforms such as TikTok and Instagram faced an identical dynamic: adversarial trends and guideline-evading content outpaced manual data pipelines, forcing a transition to zero-shot moderation architectures over static classifiers~\cite{meta2021fewshot, goldzycher2022hypothesis, liu2025mind}. Robotics now faces its own content-moderation moment—unseen objects, novel tool affordances, and unmodeled home environments form an open-world distribution that centralized collection cannot cover.

In digital AI, zero-shot capabilities matured across three distinct waves: hand-defined attribute mapping~\cite{lampert2009learning}, generative synthesis under noisy labels~\cite{felix2018multi, xian2018feature, garg2023instance}, and web-scale foundation models aligned via instruction tuning and human feedback~\cite{brown2020language, ouyang2022training, radford2021learning, wei2022finetuned}. Early vision-language-action models attempt to port these representations directly into physical control~\cite{brohan2023rt2}, yet performance degrades sharply when encountering genuinely novel tasks or environments~\cite{batra2025zero, zhou2025exploring}.

When applied directly to hardware, the digital AI toolkit falls short. However, it points to a clear pathway: a model resolves the unseen only when its latent space is rich enough to place it, handling new cases by interpolating across diverse prior representations.

\subsection{What builds a latent space}
\label{sec:rw/latent-space}
In the latent space, AI models build representations of concepts of the world. The more diverse the latent space, the more generalizable it becomes \cite{kornblith2019better, teney2021evading, fang2022data}. Representation quality is what learning ultimately depends on \cite{bengio2013representation}, zero-shot performance tracks how well the embedding space covers unseen classes \cite{xian2019zero}, and CLIP zero-shot transfer improved as its training data grew broader \cite{fang2022data}.

In physical AI, no single method yet controls a robot end to end; a working system layers several. Classical control and engineering handle sensing and low-level loops~\cite{oxe2024}. Reinforcement learning, trained largely in simulation, teaches basic motor skills such as balancing, reaching, and locomotion, effectively learning motor control from experience~\cite{schulman2017proximal, rudin2021learning}. Vision-language-action (VLA) models inherit a web-scale latent space and use it to orchestrate those skills from visual and linguistic context~\cite{brohan2023rt2, kim2024openvla, black2024pi0, nvidia2025gr00t, oxe2024}. How they connect to the body is still open: some systems map VLA outputs directly to actuators~\cite{black2024pi0}, while others have the VLA call lower-level policies, so it never commands the actuators itself, which limits risk~\cite{ahn2022saycan}. World models add a further layer: they learn a latent representation of how the world changes, so a robot can imagine several ways of completing a task, predict the outcome of each, and choose the most promising before acting~\cite{ha2018world, hafner2023mastering, murlabadia2026vjepa, zhang2026hjepa}. The architecture of physical AI is therefore far from settled.

\begin{figure}[t]
    \centering
    \includegraphics[width=.9\linewidth]{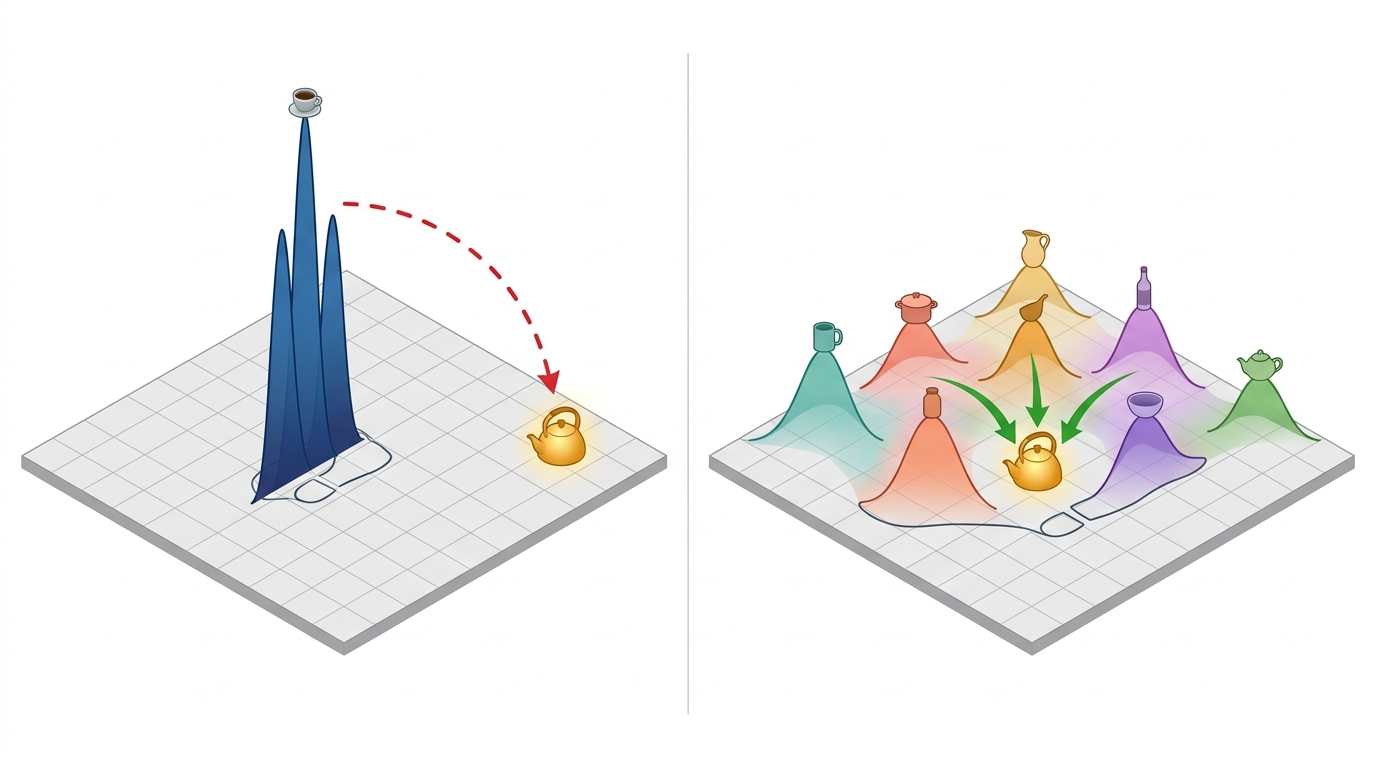}
    \caption{Volume makes each skill reliable; diversity determines whether a new task falls within reach. A robot trained on one team’s data cannot place an unfamiliar kettle in an unfamiliar kitchen. A robot trained on diverse community data reaches it by combining skills it already has.}
    \label{fig:density-diversity}
\end{figure}


Despite their progress, a systematic link between these paradigms and zero-shot generalization has yet to be established: each builds a latent space, but none addresses how the diversity of that space is sourced. Furthermore, when building foundational latent spaces, two patterns are often observed: (1) pre-training models follow scaling laws for general world contextual understanding, and (2) once pre-trained, these models require expert-level annotation for supervised fine-tuning stages \cite{kaplan2020scaling, ouyang2022training}.


\subsection{Scaling laws and Noisy-labels from crowd-source data}
\label{sec:rw/crowd-sourced}

Training data for AI has come from people in three distinct ways.

\emph{Community-driven} data is produced by people pursuing their own goals and becomes training data as a by-product, so its diversity grows with the number of people taking part~\cite{brown2020language, kopf2023openassistant}. 
\emph{Crowd-sourced} data is produced by paid workers executing tasks someone else designed, which scales and offers control but reflects its designers' choices~\cite{deng2009imagenet, ouyang2022training, mandlekar2018roboturk, agarwal2026cobalt}.
\emph{Specialized} data is produced by trained experts, in robotics mostly teleoperators on the target hardware, which matches the robot's embodiment but is the most expensive per sample and concentrated in a few teams~\cite{fernholz2026xdof, temkin2026mecka}.

Each training stage has drawn on a different form of crowd-sourcing. Pre-training, governed by scaling laws, needs volume and diversity: language models found both in text written by people all over the world as a by-product of everyday use \cite{brown2020language, raffel2020exploring}. Supervised fine-tuning needs expert, high-signal data: it came from paid annotators \cite{ouyang2022training} and, increasingly, from volunteer communities \cite{kopf2023openassistant}. Table~\ref{tab:sources} compares these sources with the workarounds robotics uses today.

These sources differ in cost, scale and diversity. Paid crowd-sourcing, from annotation micro-tasks \cite{deng2009imagenet} to paid robot teleoperation \cite{mandlekar2018roboturk}, offers control, but its cost grows with every sample and its coverage reflects its designers. Volunteer communities scale with shared motivation. Data generated as a by-product of use, as in games with a purpose \cite{vonahn2004labeling, cooper2010predicting}, scales with the number of people who want to take part.
The common response is that robotics simply lacks data volume. Paid collection yields volume without diversity. Relying on isolated teams creates expensive, narrow datasets that fail to capture the long-tail reality of physical environments. Language models, by contrast, sourced each stage from the crowd. A further barrier is hardware. Robots are expensive for companies to lend to collectors and expensive for annotators to buy, so collection stays concentrated in a few well-funded teams. Workarounds that avoid robots altogether, such as egocentric human video, lower the cost but introduce an embodiment gap: human hands and viewpoints differ from a robot’s, which limits how well the data transfers \cite{grauman2022ego4d, kareer2024egomimic}.
Robotics relies largely on paid collection for both stages. Companies have raised substantial capital~\cite{figure2026index}, with teleoperation and egocentric data collection a major cost \cite{fernholz2026xdof, temkin2026mecka, afp2026india}, yet their models still need new demonstrations for each new task \cite{black2024pi0, zhou2025exploring}. A notable exception is the LeRobot community, whose contributed datasets on the low-cost SO-100 arm were used to pre-train SmolVLA \cite{shukor2025smolvla, cadene2024lerobot}. These open tools, however, are built for research speed: they evolve quickly, often without backward compatibility, and favor rapid experimentation over the stability needed to build reliable applications. Their community data is a by-product of writing papers, not of solving everyday problems, so it stays concentrated on the tasks researchers choose to study. Our approach differs in where contributions come from: a heterogeneous multi-environment collection by developers building applications on one stable, standard robot and platform, whose data reflects the problems they actually need to solve.

\begin{table*}[!t]
\centering
\small
\begin{tabular*}{\textwidth}{@{\extracolsep{\fill}}lcccc}
\toprule
Source & Cost / hour & Diversity & Embodiment & Scales with \\
\midrule
Paid teleoperation         & High     & Low      & Match   & Budget \\
Egocentric human video     & Very low & High     & Gap     & Budget \\
Simulation only            & Low      & Designed & Sim gap & Compute \\
Research community         & Low      & Medium   & Match   & Papers \\
Developer community (ours) & Low      & High     & Match   & Users \\
\bottomrule
\end{tabular*}
\caption{Sources of robot training data. Diversity is how widely a source covers situations in $P_{\text{world}}$; embodiment is whether the data comes from the robot that will use it. Only a developer community scales with the number of people solving their own problems on the target robot.}
\label{tab:sources}
\end{table*}

Heterogeneous multi-environment collection is never clean. In classification, learning with noisy labels has developed methods that select likely-clean samples during training~\cite{han2018coteaching, li2020dividemix}, model noise that depends on each instance rather than on its class alone~\cite{xia2020part, garg2023instance}, and estimate how much noise a dataset contains~\cite{garg2025aeon, garg2024eccv}. In robot learning, the quality of demonstrations matters as much as their quantity: policies trained on demonstrations from operators of mixed skill perform markedly worse than those trained on proficient operators alone~\cite{mandlekar2021matters}. Imitation methods have so far handled this mainly by filtering, discarding low-quality demonstrations~\cite{mandlekar2021matters, agarwal2026cobalt}. Filtering protects performance but discards diversity, and for zero-shot generalization diversity is the scarcer resource. We therefore treat imperfect demonstrations as noisy labels to be weighted rather than discarded (Section~\ref{sec:method/latent}).


\subsection{Trust and Safety in Physical AI}
\label{sec:rw/trust-and-safety}

Most robot safety work today addresses collision avoidance, joint limits, and grasp force~\cite{zacharaki_safety_2020, brunke_semantically_2025, hu_vlsa_2025}, and treats broader safe behavior as a layer added after pre-training: a foundation model is trained to act, then fine-tuned on examples of what unsafe means~\cite{zhang_safevla_2025, jindal_can_2025}. This leaves two questions open. First, how can a robot avoid unsafe behavior in situations for which no data can ethically or practically be collected? Second, why is safety absent from the stage where the latent space is formed? Both are zero-shot problems, and the same foundations address them. Rare or dangerous situations can be generated rather than staged, so the robot learns them without putting anyone at risk~\cite{yu2023scaling,jindal_can_2025,felix2020augmentation}. The robot should also recognize when it has left familiar ground: building on generalized zero-shot learning with domain classification, which separates seen from unseen inputs~\cite{felix2019domain,felix2020augmentation}, and on estimating out-of-distribution noise~\cite{garg2025aeon}, it can detect when a situation falls in $\mathcal{S}_{\text{unseen}}$ and stop or hand control to a human through teleoperation. 

\section{Building a Zero-Shot Foundational Latent Space}
\label{sec:method}


\subsection{Problem Formulation}
\label{sec:method/formulation}

A deployed robot meets a stream of situations, each a combination of its own state, the task it is asked to do, and the world around it, so a familiar task in an unfamiliar kitchen is already a new situation. Because these elements combine, the space of situations is vast and long-tailed: most situations are individually rare, yet together they make up a large share of what a robot meets in practice~\cite{liu2019large, feldman2020does}. We define zero-shot generalization as the robot's success on situations its training data never covered, and it depends on two properties of that data. Volume makes the robot reliable on the situations the data covers; diversity decides which situations are covered at all (Fig.~\ref{fig:density-diversity}). Today's collection supplies volume but little diversity, because a few teams collect in the settings they can foresee and access. Each new contributor brings a setting no one else has and reaches situations the others cannot, and the rarer a situation, the more contributors it takes to reach it. Diversity is therefore a requirement alongside volume, and it scales with the number of contributors. A formal treatment is given in Supplementary Section~\ref{supp:formulation}.


\subsection{Zero-Shot Generalization Through a Shared Latent Space}
\label{sec:method/latent}

Zero-shot capability benefits from a mixture of models to meet in a common representation~\cite{radford2019language, mustafa2022multimodal,girdhar2023imagebind, radford2021learning}. We therefore build a shared latent space for robot actions. The space is not tied to a single architecture: policies learned through reinforcement learning, vision-language-action models, and world models can all draw on it, whether they are realized as one model or several. We build this space from complementary principles.

\noindent\textbf{Alignment across modalities.} Multi-modal alignment places vision, language, and the robot's own state and actions in one space, so a task described in words can connect to actions never demonstrated in that exact form~\cite{felix2020dicta, felix2019ensemble, radford2021learning, mustafa2022multimodal, brohan2023rt2}. Cross-modal learning then carries knowledge between modalities, for example from a task description to an action, or from one contributor's setting to others~\cite{felix2018multi, felix2021dicta}.

\noindent\textbf{Augmentation.} Generative models synthesize variations of contributed data, such as new objects, layouts, and conditions, filling the gaps between what contributors recorded~\cite{felix2018multi, felix2020augmentation, xian2018feature, yu2023scaling}.

\noindent\textbf{Learning from imperfect data and uncertainty.} Contributions vary in quality. Noisy-label learning lets the space absorb imperfect demonstrations without discarding the diversity they bring~\cite{garg2023instance, garg2024eccv, garg2025pass}. Bayesian uncertainty learning separates noise in the data from gaps in the model's knowledge~\cite{gal2016dropout, kendall2017uncertainties}.


\noindent\textbf{Safety from pre-training onward.} Physical AI needs the same capability developed for digital AI, at higher stakes: an error is an action in the world, not a digital asset that can be taken down. We carry these tools into the shared latent space in three ways. First, rare or dangerous situations are synthetically generated rather than staged and enter training alongside contributed data~\cite{yu2023scaling}. Second, seen/unseen detection~\cite{felix2019domain} and out-of-distribution noise estimation~\cite{garg2025aeon} operate on the same representation the policy acts from, so when a situation falls in $\mathcal{S}_{\text{unseen}}$ the robot stops or hands control to a human through teleoperation. Third, high-level models call validated lower-level policies rather than commanding low level components directly, a layer the shared latent space can inform but not bypass~\cite{varley_embodied_2024, li_safe_2024}. Placing these signals in the shared latent space, rather than in a separate filter, makes safety a property of what the robot has learned.

Every layer of this stack is developed first in simulation: a policy that does not perform well there is never deployed on hardware. Yet a policy that succeeds in simulation often fails on the real robot, because real physics, sensing, wear, and noise differ from what the simulator models~\cite{zhao2020sim, openai2019rubik}, as illustrated on Figure~\ref{fig:platform}. This sim-to-real gap is not a weakness of zero-shot methods themselves; it affects any learning-based robot. It does, however, block zero-shot generalization in practice: a robot cannot solve an unseen task in the real world if what it learned does not survive the move from simulation.

\subsection{Closing the Sim-to-Real Gap with Real2Sim2Real}
\label{sec:method/real2sim2real}
Robots improve fastest when learning moves freely between simulation and the physical world: real deployments calibrate the simulation, and simulation lets new behaviors be trained and tested before they reach the real robot~\cite{chebotar2019closing, torne2024reconciling}. We build this loop on a hardware platform fully integrated with a software platform for task training, designed for fast iteration between the two. 

\begin{figure}[t]
  \centering
  \includegraphics[width=0.9\linewidth]{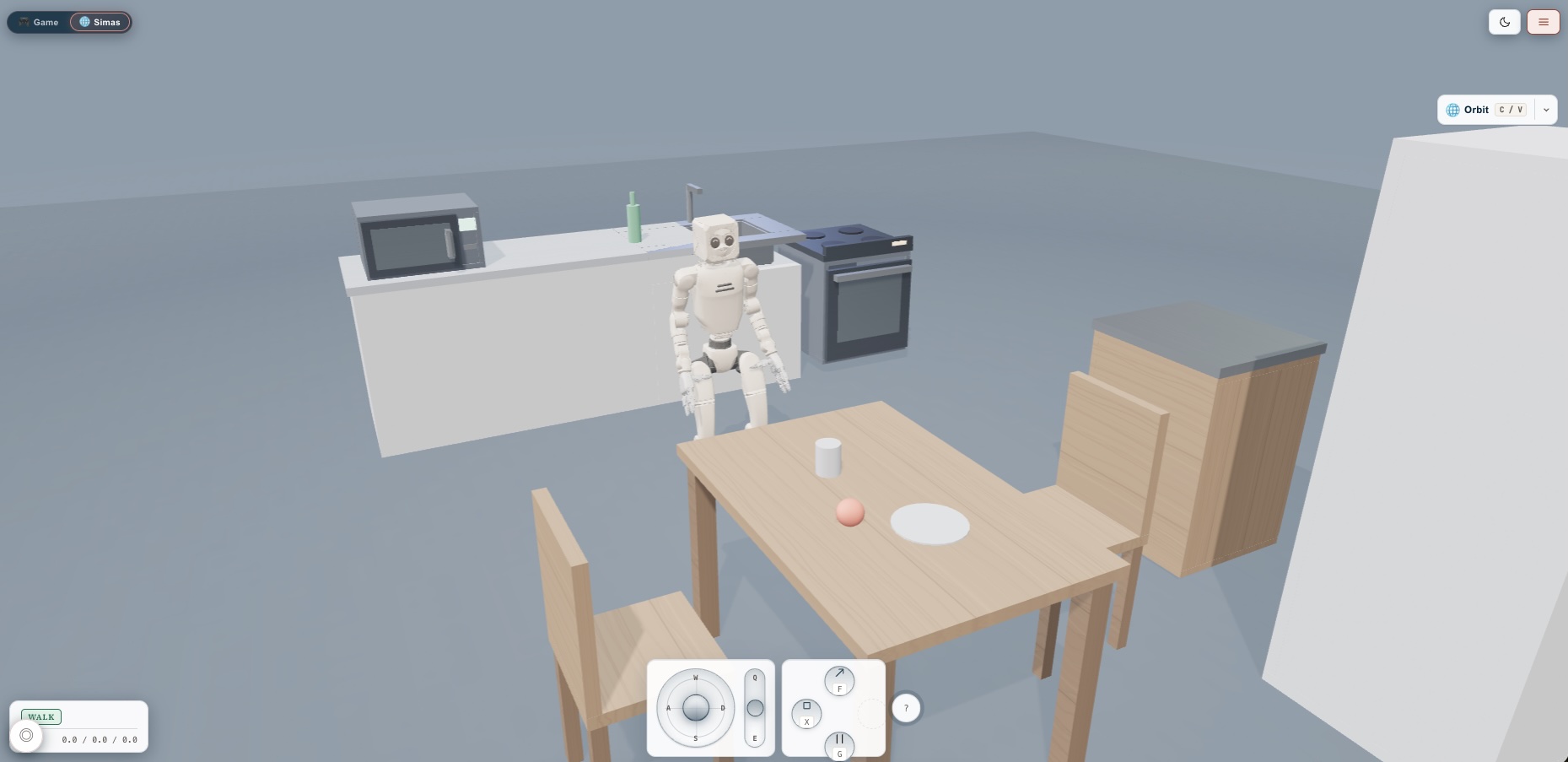}
  \caption{The training platform in an online environment: teaching by demonstration, designing policies, and setting challenges.}
  \label{fig:platform}
\end{figure}

\begin{figure}[t]
  \centering
  \includegraphics[width=0.75\linewidth]{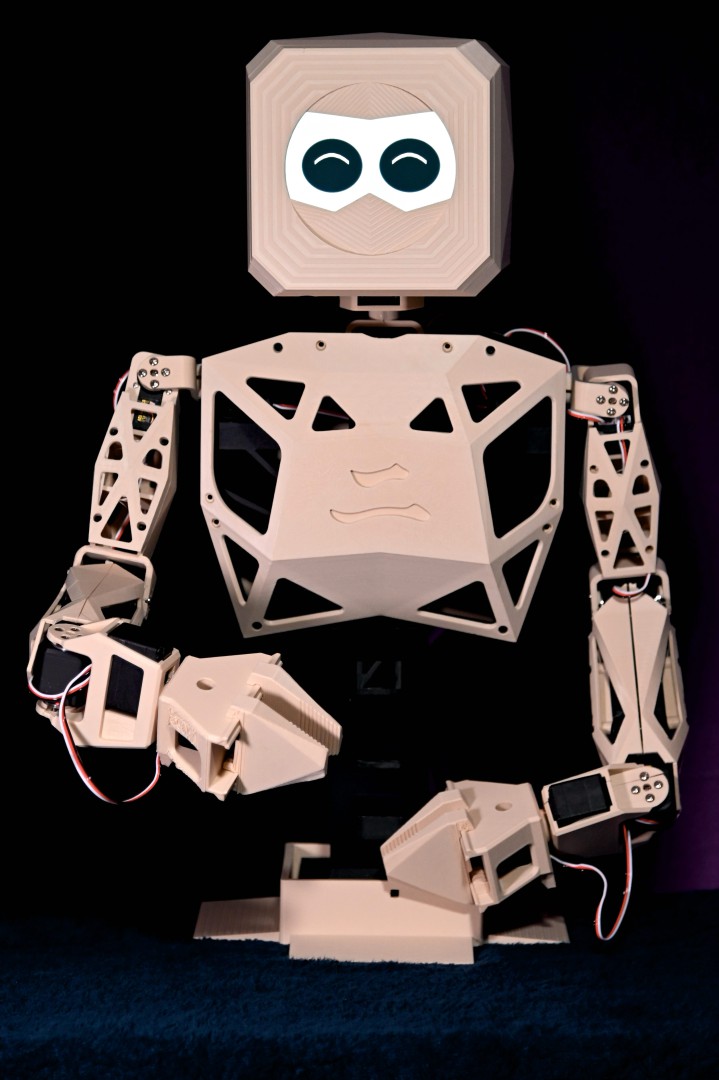}
  \caption{Origami Dual-Arm: an upper-body platform with 14 DoF that learns by record-and-replay and has a browser twin for training before deployment.}
  \label{fig:origami}
\end{figure}

\begin{figure}[t]
  \centering
  \includegraphics[width=0.75\linewidth]{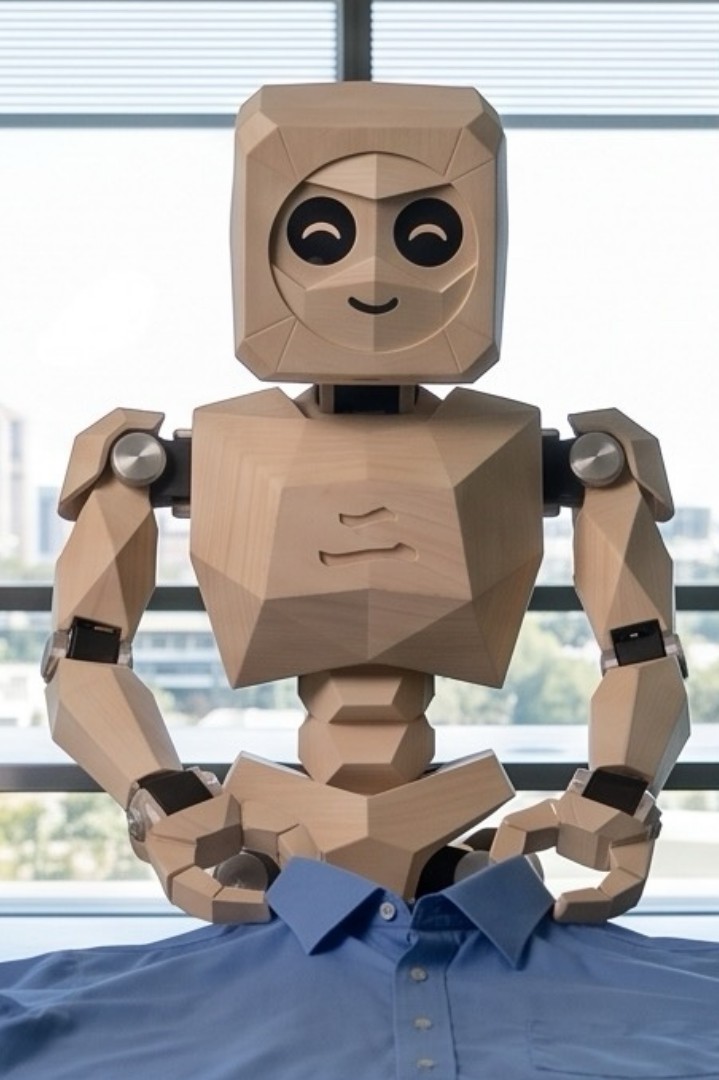}
  \caption{The vision for Origami: a 27-DoF bipedal humanoid with multi-view sensing and on-board AI compute. Specifications are design targets.}
  \label{fig:origami-future}
\end{figure}

Two components of this platform are in progress. First, a training platform where developers teach the robot through demonstration, design their own policies, and set challenges that test what the robot can do (Figure~\ref{fig:platform}). Second, Origami, an approachable robot designed to be safe and to let each developer progress at their own pace (Figure~\ref{fig:origami}, and Figure~\ref{fig:origami-future}).

We call a robot \emph{approachable} when an individual developer can own it, start building on it, and operate it safely without the
resources of a laboratory: it is affordable to an individual, usable from first session without specialist setup, designed for safe operation around people, and stable enough that what is built on it keeps working as the platform evolves.

\begin{table*}[t]
  \centering
  \small
  \begin{tabular}{lll}
    \toprule
     & \textbf{Origami Dual-Arm} & \textbf{Origami Premium} \\
    \midrule
    DoF          & 14 (neck 2, arms 6 each)          & 27 (neck 3, arms 7, legs 5 each) \\
    Size / mass  & 57\,cm, 2.8\,kg (upper body)      & $\sim$110\,cm, $\approx$18--25\,kg \\
    Manipulation & 2$\times$6-DoF arms, 1-DoF grippers & 2$\times$7-DoF arms, 1-DoF grippers, 600\,g payload \\
    Locomotion   & Stationary                         & Bipedal walking \\
    Sensing      & Camera, IMU, Joint position and load            & 3 wide cameras (head, chest, hand), IMU, \\
                 &                                    & per-actuator position, current, temperature \\
    Actuation    & Smart servos                       & Quasi-direct-drive \\
    Compute      & Single-board computer              & AI compute module + real-time controller \\
    Interaction  & 5$''$ touch face                   & 5$''$ touch face, 4-mic array, speakers \\
    Power       & 12\,V tethered                     & Battery, e-stop, current/temperature monitoring \\
    Software     & SDK, Record \& replay, browser twin     & + teleoperation, learned policies\\
    \bottomrule
  \end{tabular}
  \caption{Origami specifications. The Premium version includes every feature of the dual-arm version and adds the capabilities listed. Humanoid values are design targets and may change.}
  \label{tab:origami-specs}
\end{table*}

Developers therefore run the same robot on the same platform; each deployment shows where simulation and reality differ, and each correction improves the digital twin for everyone. Neither simulation nor reality is a perfect label. Building on instance-dependent noisy-label learning~\cite{garg2023instance}, we treat each as a noisy teacher of the other, so robots learn what holds in both.

Origami is designed for a gap in today's hardware (Figure~\ref{fig:origami}), with specs described in Table~\ref{tab:origami-specs}. Programmable research platforms remain expensive: even systems built explicitly to be low-cost sit at \$18--32k~\cite{zhao2023learning, fu2024mobile}, and commercial mobile manipulators such as the PR2 and TIAGo can exceed \$200k~\cite{fu2024mobile}. This limits most laboratories to a handful of robots and puts them out of reach for individual developers, especially outside well-funded institutions. At the other end, open printable arms cost under \$1{,}000 and have built an active community~\cite{cadene2024lerobot, shukor2025smolvla}. These projects, however, tend to focus on one layer of the stack: developers work directly at the actuator level, without a stable control layer, safety protocols, or reliable compatibility with standard robotics middleware~\cite{macenski2022ros2}, and a policy trained for one task rarely carries over to the next~\cite{zhou2025exploring}. Origami targets the space between the two: a robot approachable enough for an individual developer, and stable enough to build applications on.


\subsection{Latent Space via High-Entropy Data Sourcing}
\label{sec:method/community}
Developers already contribute freely when the work matters to them: Wikipedia~\cite{wikipedia}, OpenStreetMap~\cite{haklay2008openstreetmap}, Hugging Face~\cite{huggingfacehub}, and many other initiatives. Our approach leverages that energy with a shared robot and a shared platform, with paid annotators adding a quality layer whenever expert labels are needed.

The demand for robotic engineers is approaching an inflection point. As robotics shifts from single-task industrial arms into multi-modal foundation models and humanoids, the field faces a strong demand for talent~\cite{hamblen2026openai}. In past decades, US employment in computer systems design grew roughly $6\times$ between 1990 and 2023 (Fig. 6)~\cite{fred2026}, and global expansion mirrors this trend. However, the World Economic Forum projects structural labor churn of 22\% through 2030, with AI/ML specialists and robotics engineers ranking among the fastest-growing job titles globally~\cite{wef2025jobs, wef2023jobs}. Building a developer-first platform provides the shared infrastructure and hardware access where this next generation of physical AI builders learns, iterates, and deploys.

\begin{figure}[t]
  \centering
  \includegraphics[width=\linewidth]{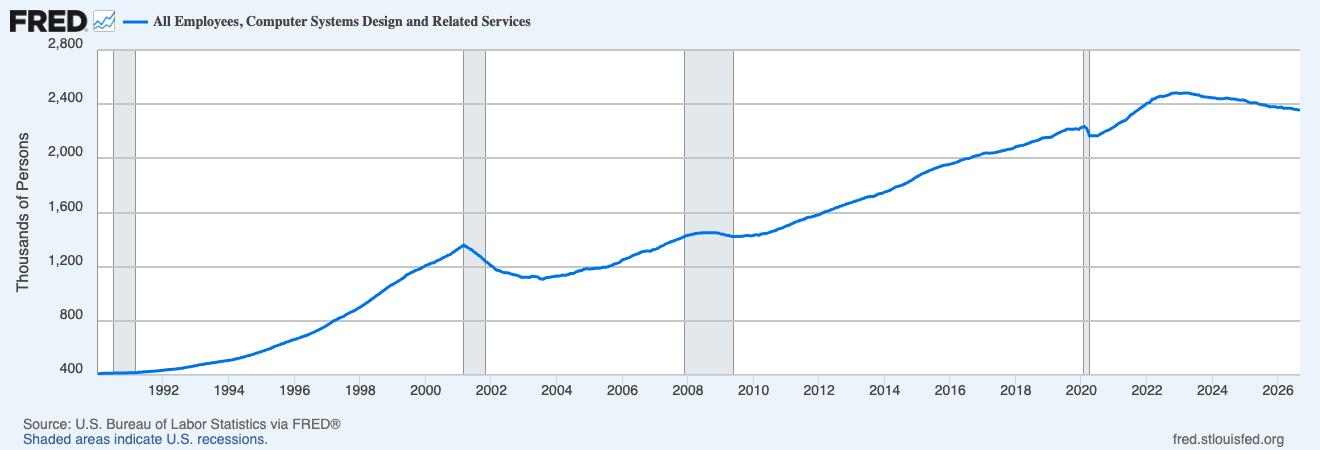}
  \caption{Growth of all employees in computer system designs and related services from 1990 to 2023 in the US.}
  \label{fig:engineers_growth}
\end{figure}

This changes the cost of data. Paid robot data collection costs roughly \$50--\$200 per hour~\cite{rcsv2026cost, figure2026index}. On our proposed platform, developers contribute data as a by-product of solving their own problems, such as building an application for a workplace or automating a task at home. The cost of a contributed hour is therefore set by acquisition and retention rather than by labor: the cost of acquiring one developer divided by the usable hours that developer contributes over their time on the platform. Taking \$3.50 per developer, at the upper end of user-acquisition costs for online games~\cite{businessofapps2024cpi}, a single usable hour brings the cost to about \$3.50 per hour, 14--57$\times$ below paid collection, and every further hour a developer contributes lowers it again.


\textbf{From platform to training data.} Contributions are captured through the platform's task interface. Each task states a goal, such as stacking, sorting, or handing over an object, but not how to achieve it: the developer decides the strategy, the grasp, and the sequence. Different contributors therefore solve the same task in different ways, which adds diversity to the actions themselves, not only to the objects and settings. Challenges set by one developer become tasks for others, so the task distribution grows with the community rather than with our team's foresight. Every session is recorded with its outcome, including partial and failed attempts, which are informative about where policies break. Before contributed data enters training, it passes through three stages: automatic checks for corrupted or physically implausible trajectories, noise-aware weighting of the remaining demonstrations (Section~\ref{sec:method/latent}), and expert review for a sampled subset. Finally, the same infrastructure built for community-driven data also serves crowd-sourced annotation and specialized teleoperation.
\section{Hypothesis and Evaluation Protocol}
\label{sec:hypothesis}

\subsection{Hypothesis}
\label{sec:hyphothesis/hypothesis}

Each component needed for a foundational latent space for zero-shot robotics has been demonstrated separately: affordable robots and open
tooling have lowered the barrier to building with robots~\cite{cadene2024lerobot, macenski2022ros2}; generative zero-shot learning and noisy-label learning provide the methods~\cite{felix2018multi, garg2023instance}; large open robot datasets show that pooled, diverse data
transfers~\cite{oxe2024}; and communities have built training data before~\cite{kopf2023openassistant, agarwal2026cobalt,shukor2025smolvla}. What has not yet been done is to bring these components together on one platform and test whether contributor diversity, rather than data volume alone, drives zero-shot success on real robots. We state this as our central hypothesis:

\begin{hypothesis}[Diversity over volume]
\label{hyp:diversity}
At a fixed number of training samples, real-robot zero-shot success
$G_{\text{zs}}$ increases with the number of distinct contributors,
environments and objects in the training set, and increases more than
adding the same number of samples from existing contributors.
\end{hypothesis}

\begin{figure}[t]
  \centering
   \includegraphics[width=0.8\linewidth]{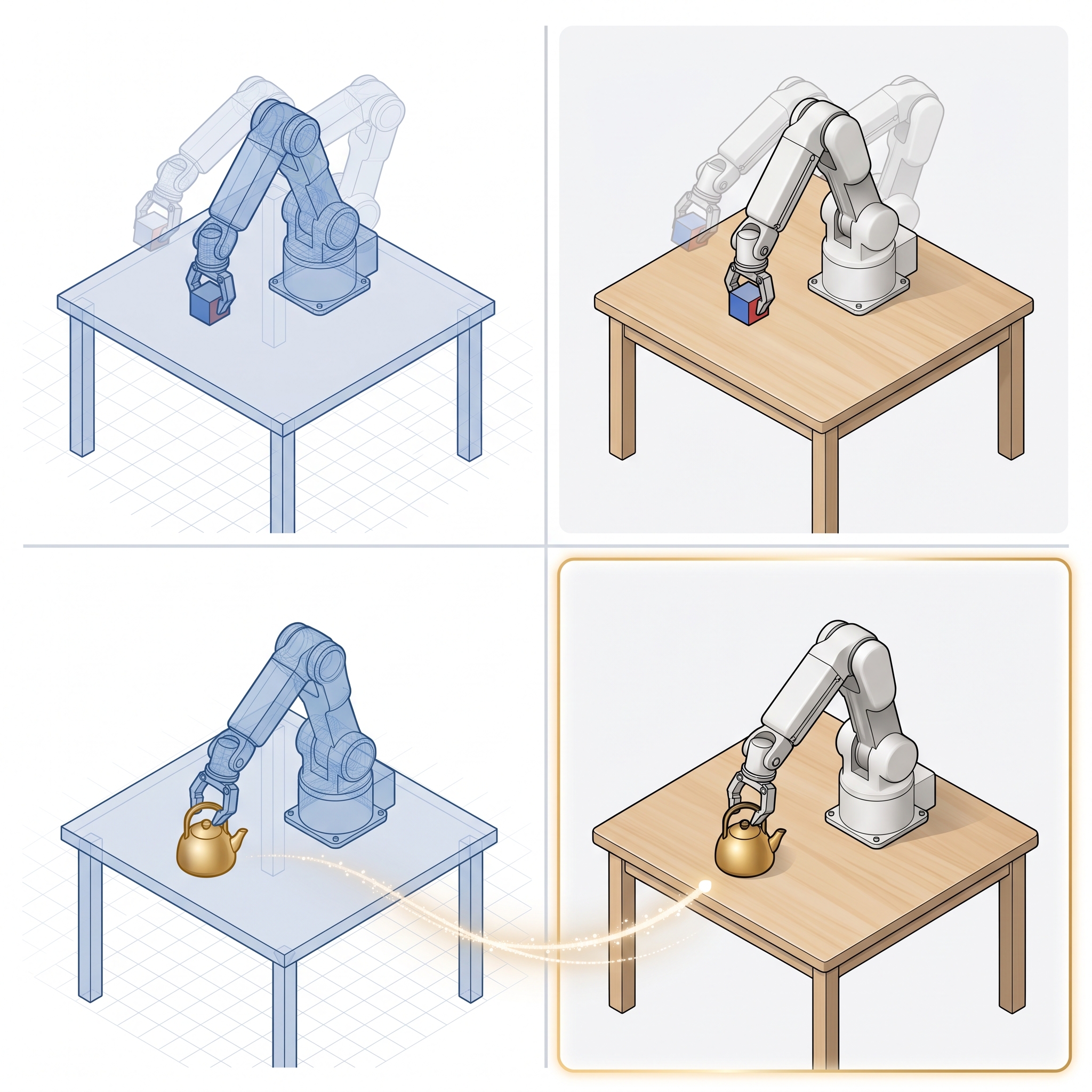}
   \caption{Columns: simulation (left) and real robot (right). Rows: seen task (top) and unseen task (bottom). The target is the highlighted quadrant: an unseen task solved on the real robot using knowledge built largely in simulation.}
   \label{fig:measurement-hero}
\end{figure}

\noindent The rest of this section sets out how we test it. We measure progress along two axes. Zero-shot generalization is the goal: success on tasks, objects and environments the robot never trained on. Sim-to-real transfer is the enabler: how much of a policy’s performance in simulation carries over to the real robot on the same task. The strongest result sits where the two meet: an unseen task, solved on the real robots, as depicted in Figure~\ref{fig:measurement-hero}.

We are establishing a baseline on the Origami platform, in simulation and on the physical robot, using a benchmark of tasks defined by our team and extended through decentralized, long-tail data acquisition. Each benchmark task is held out from training for at least one evaluation, so seen and unseen performance can be compared directly. Our central hypothesis is that zero-shot success on the real robot increases with the diversity of contributed data, measured by the number of distinct contributors, environments and objects in the training set. Support for Hypothesis~\ref{hyp:diversity} already exists in robot manipulation, where generalization follows a power law in environment and object diversity~\cite{lin2025data}. We further expect sim-to-real transfer to improve as the contributor base grows, since each real deployment exposes a new gap between simulation and reality (Section~\ref{sec:method/real2sim2real}). We will report both measures as the developer base expands.


\subsection{Evaluation protocol}
\label{sec:hyphothesis/evaluation}

We split the benchmark along the three components of a situation $s=(x,\tau,w)$, giving four held-out conditions of increasing difficulty: an unseen object in a seen environment, an unseen environment for a seen task, an unseen task in a seen environment, and an unseen combination of task and environment. For each condition we report success rate in simulation and on the real robot. We estimate $G_{\text{zs}}$ as the average success over held-out conditions, weighted by how often each appears in a sample of deployment requests from the distributed real-world domain sampling. We measure sim-to-real transfer as the ratio of real to simulated success on the same task. To test the central hypothesis, we train on nested subsets of contributed data that hold the total number of samples fixed while varying the number of contributors, environments, and objects. If diversity matters more than volume, $G_{\text{zs}}$ should rise with the number of contributors at a fixed data budget, following the trend reported for manipulation~\cite{lin2025data}, while adding samples from the same contributors should yield smaller gains. A complete formal treatment is provided in Supplementary~\ref{supp:formulation}.
\section{Limitations and Open Questions}
\label{sec:limitations}

Our proposal rests on assumptions that remain to be tested. \textbf{Consent and data rights:} contributed sessions must be collected with explicit consent and clear terms on how they are used, retained, and deleted; trust is a precondition for the community, not a by-product of it. \textbf{Quality:} decentralized, long-tail data acquisition varies in skill and care of the developers; noisy-label learning reduces but does not remove the need for validation, and expert annotation remains necessary for some stages. \textbf{Simulation bias:} most contributions will begin in simulation, so gains in zero-shot success may not transfer until \textit{Real2Sim2Real} closes enough of the gap. \textbf{Coverage bias:} contributors in a decentralized model remain a subset of the global distribution; contributors may cluster in certain regions, languages, and settings, leaving parts of the world uncovered. \textbf{Acquisition cost:} no acquisition data yet exists for developer-contributed robot data, so our estimate is drawn from user acquisition in online games, taken at the upper end of the reported range~\cite{businessofapps2024cpi}. \textbf{Retention:} the lowest cost per hour depends on developers staying active. Where retention falls short, the same platform supports paid crowd-sourced teleoperation, which has already been shown to run on everyday devices at scale~\cite{mandlekar2018roboturk, agarwal2026cobalt}; this raises the cost per hour but keeps it below specialized collection, at the price of the diversity that self-directed contributors bring.




\section{Conclusions}
\label{sec:conclusions}

This work establishes a framework for physical zero-shot generalization grounded in two pillars: a decentralized data-acquisition pipeline deployed on a standardized, low-cost robot, and a unified latent space that converts heterogeneous real-world contributions into transferable motor skills. The space combines multi-modal alignment, generative augmentation and learning from imperfect data. Together, they let a robot handle situations it was never trained on.


Generalization is what turns a robot from a single-task machine into an economic asset. We discussed that what robotics lacks is not volume but diversity: it has begun to scale the amount of data, but not the range of situations that data covers. Diversity grows with the number of independent contributors rather than with the size of a few paid teams. Our platform also serves a second function: developers who contribute data acquire hands-on experience with physical AI systems, so the community that supplies diverse data simultaneously develops the workforce the field requires.


Finally, a growing wave of developers wants to learn and build with robots. This decentralized base provides the most sustainable source of data diversity that zero-shot generalization requires. Without it, robotics risks remaining confined to single-task machines while its economic and social potential stays out of reach.

{
    \small
    \bibliographystyle{ieeenat_fullname}
    \bibliography{main}
}

\clearpage
\setcounter{page}{1}
\maketitlesupplementary

\section{Problem Formulation}
\label{supp:formulation}
A robot faces a situation $s = (x, \tau, w)$: its state $x$ (joint positions, velocities, forces), a task $\tau$, and a world state $w$ (objects, environment, setting). At deployment, situations follow a distribution $P_{\text{world}}$ over $\mathcal{S} = \mathcal{X} \times \mathcal{T} \times \mathcal{W}$, shaped by where robots operate: across homes, countries, cultures, and tasks.

A policy $\pi$ is trained on data drawn from a distribution $P_{\text{train}}$ and is then deployed unchanged, with no situation-specific adaptation. Let $R \in [0, 1]$ be the probability that $\pi$ completes task $\tau$ from state $x$ in world $w$. Expected task resolution in deployment is
\begin{equation}
    G = \mathbb{E}_{s \sim P_{\text{world}}} [R].
\end{equation}
Let $\mathcal{S}_{\text{seen}}$ be the situations covered by the training data, and $\mathcal{S}_{\text{unseen}} = \mathcal{S} \setminus \mathcal{S}_{\text{seen}}$. Then
\begin{equation}
    G = p_{\text{seen}} \mathbb{E}[R \mid \mathcal{S}_{\text{seen}}] + p_{\text{unseen}} \mathbb{E}[R \mid \mathcal{S}_{\text{unseen}}],
\end{equation}
where $p_{\text{unseen}} = P_{\text{world}}(\mathcal{S}_{\text{unseen}})$. We write $G_{\text{seen}} = \mathbb{E}[R \mid S_{\text{seen}}]$ for success on situations covered by training, and call the second expectation zero-shot
generalization, $G_{\text{zs}} = \mathbb{E}[R \mid S_{\text{unseen}}]$. Most robot benchmarks report $G_{\text{seen}}$; deployment is governed largely by $G_{\text{zs}}$.

The problem is hard for three reasons. First, $\mathcal{S}$ is combinatorial, so a familiar task in an unfamiliar world is already a new situation. Second, $P_{\text{world}}$ is long-tailed, so individually rare situations together carry substantial probability~\cite{liu2019large, feldman2020does}; $p_{\text{unseen}}$ therefore stays large, and the term $p_{\text{unseen}}\, G_{\text{zs}}$ dominates $G$.. Third, $P_{\text{train}}$ is designed by whoever collects the data. When a few teams collect it, $P_{\text{train}}$ concentrates on the situations they foresee, and more samples from the same distribution cover $\mathcal{S}_{\text{unseen}}$ only slowly as depicted in Fig. \ref{fig:density-diversity}. 

When data comes instead from $K$ contributors, each operating in their own setting,
\begin{equation}
    P_{\text{train}} = \frac{1}{K} \sum_{k=1}^{K} P_k.
\end{equation}

Each new contributor can add support where $P_{\text{world}}$ has mass but no existing contributor reaches. Zero-shot generalization is therefore limited by how much of $P_{\text{world}}$ the training distribution covers, not by how many samples it contains.

Because the support of a mixture is the union of its components' supports, $\operatorname{supp}(P_{\text{train}}) = \bigcup_{k=1}^{K} \operatorname{supp}(P_k)$, coverage can grow in two ways. An existing contributor can collect more, extending $\mathcal{S}_{\text{seen}}$ into new situations, but their reach is shaped by one setting and one set of habits, so each additional sample increasingly revisits familiar ground. A new contributor instead brings a setting that no existing contributor reaches. If contributors' settings are drawn independently, and each reaches a situation $s$ with probability $q(s)$, the expected unseen mass after $K$ contributors is

\begin{equation}
  \mathbb{E}\big[p_{\text{unseen}}(K)\big] = \int_{\mathcal{S}}
  P_{\text{world}}(s)\,\big(1 - q(s)\big)^{K}\, ds,
  \label{eq:unseen}
\end{equation}
which decreases monotonically in $K$. The decrease is slowest exactly where $q(s)$ is small: the rare, long-tailed situations that a few collection teams are least likely to foresee. Volume from the same contributors helps, but diversity of contributors is what drives this term toward zero.

\end{document}